# SENSEMAKING IN NOVEL ENVIRONMENTS: HOW HUMAN COGNITION CAN INFORM ARTIFICIAL AGENTS

Robert E. Patterson*, Regina Buccello-Stout*, Mary E. Frame†, Anna M. Maresca†, Justin Nelson*, Barbara Acker-Mills†, Erica Curtis†, Jared Culbertson‡, Kevin Schmidt‡, Scott Clouse‡, and Steve Rogers‡
*Air Force Research Laboratory
†Parallax Advanced Research
‡Autonomy Capability Team (ACT3) Wright-Patterson AFB

## I. ABSTRACT

One of the most vital cognitive skills to possess is the ability to make sense of objects, events, and situations in the world. In the current paper, we offer an approach for creating artificially-intelligent agents with the capacity for sensemaking in novel environments. Objectives: to present several key ideas: (1) a novel unified conceptual framework for sensemaking (which includes the existence of sign relations embedded within and across frames); (2) interaction among various content-addressable, distributed-knowledge structures via shared attributes (whose net response would represent a synthesized object, event, or situation serving as a sign for sensemaking in a novel environment). Findings: we suggest that attributes across memories can be shared and recombined in novel ways to create synthesized signs, which can denote certain outcomes in novel environments (i.e., sensemaking).

## II. INTRODUCTION

One of the most important cognitive skills to possess is the ability to make sense of objects, events, and situations in the world. The capability for sensemaking is critical when comprehending everyday situations, such as when individuals engage in social interaction with coworkers, make critical decisions at work, navigate through traffic during rush hour, or avoid injury in dangerous situations. It is likely that all other cognitive processes, such as perception, attention, and memory, act in the service of sensemaking[1]. Indeed, sensemaking is important for biology as a whole [2].

In the current paper, we offer an approach for creating sensemaking in novel environments informed by human cognition. We first define the term 'sensemaking' by presenting a unified framework for the concept. Moreover, we also consider novel environments in which sensemaking can be particularly challenging. In novel environments, context is confusing—which elements in a scene should be attended to and which elements can be ignored? In novel environments, cause and effect may be unknown—which variables are related and which variables are independent? Given the importance of sensemaking in novel environments, the present paper offers an approach for creating such sensemaking as informed by human cognition.

In the field of artificial intelligence (AI), a number of recent breakthrough developments have led to significant improvements on academic benchmarks as well as numerous commercial successes. For example, Krizhevsky, Sutskever and Hinton [3] demonstrated substantial improvement on the 2010 ImageNet Large Scale Visual Recognition Challenge [4] with the development of AlexNet, a deep neural network, or DNN ([5], [6]), trained using supervised learning. The field of AI has also been successful when coupling reinforcement learning with DNNs. Silver, Huang, Maddison, Guez et al. [12] combined supervised learning of hundreds of thousands of human expert games with reinforcement learning during millions of games of self-play, to create a DNN that beat the human world champion at the game of Go [11]). More recently, significant advances have enabled these deep reinforcement learning models to be used in many novel applications, including unmanned aircraft [16]. Recent years have also seen an explosion in advances in generative AI capabilities, i.e., the development of models trained to generate text, imagery, videos or other media, often based on text prompts. The continuing development of latent diffusion models (e.g., [19]) have led to a multitude of image generation tools now available to the public. Perhaps more important was the development of the transformer architecture [22], which is now the dominant architecture for many deep learning applications due to its intrinsic parallelizability and utility with self-supervised training in sequence modeling. This has led to the surprising effectiveness of large language models (LLMs) such as GPT-4 [23], which can incorporate human feedback [24] and be incorporated into more sophisticated reasoning systems that use mechanisms such as reflection [25].

The algorithms described above primarily achieved success in narrowly-defined problem domains such as a particular reinforcement learning environment (though LLMs have been shown to have utility across a wide variety of applications provided that a natural language interface can be developed for the domain; see e.g., [26], for both capabilities and limitations of GPT-4). High levels of supervised learning typically occur only with large amounts of training stimuli (e.g., thousands of training samples) and a close match between the training and test stimuli [27]. However, when training stimuli are limited, or when test stimuli differ significantly from training stimuli, machine-learning systems perform poorly—the learning is superficial and brittle ([28]; [29]; [30]).

For example, subtle changes in an image can cause a machine-learning algorithm to mislabel a lion as a library [31], misidentify real- world stop signs as speed-limit signs [32], confuse 3D-printed turtles for rifles [33], mislabel static white noise as a lion, or misclassify yellow-and-black patterns of stripes as school buses [34]. These are important considerations for the deployment of AI solutions to solve real-world problems (e.g., adversarial machine learning; [35]).

Importantly, the algorithms described above did not entail sensemaking in the way we will describe below, but rather were constrained to either simple responses (e.g., classification models) or simple reasoning (e.g., policy-following in reinforcement learning or feedforward execution in a transformer network). And even with simple responses, there were performance problems when the testing was accomplished using stimuli that were not encountered during training—that is, when the test occurred in a novel environment. Accordingly, it would seem that the creation of artificial agents who can make sense of novel environments—the focus of the current paper—may be a very difficult proposition to execute ([36]; [37]; [38]). Nonetheless, the present paper attempts to make some progress toward that goal by presenting a number of key ideas in human cognition and considering how those ideas might inform the design of artificial agents who can make sense in novel environments.

These key ideas, discussed throughout this paper, are the following: (1) a unified framework for sensemaking; (2) pattern synthesis and memory recombination; and (3) distributed representations.

## III. UNIFIED FRAMEWORK FOR SENSEMAKING

The capability to make sense of our surrounding environment seemingly requires sophisticated cognitive processing. As Fuster [39] notes, in humans, all cognitive functions are interdependent: language depends on perception, attention, and memory, and intelligence depends on perception, attention, memory, language, and reasoning. The neural foundation of these cognitive functions is amazingly complex. The human brain typically contains slightly less than 100 billion neurons and has approximately 150 trillion synapses [40]. This complexity entails systems containing thousands of non-linear feedback loops, both positive (self- reinforcing) and negative (self-correcting) feedback, coupled to one another with multiple time delays, non-linearities, and accumulations, which can generate very complex endogenous behavior in a system ([41]; [42]). Such complex behavior implies that it will be very challenging to model human sensemaking. Yet before covering the topic of sensemaking, we will need to first discuss dual processing.

### A. *Dual Processing*

The contemporary literature shows that human decision making is mediated by two distinct sets of cognitive processes or systems, called intuitive cognition (‘system 1’) and analytical cognition (‘system 2’) ([43], [44],[45]; [46]; [47]; [48]; [49]; [50]; [51]; [52]; [53]). A quick and brief overview of the literature shows that intuitive cognition—which is unconscious situational pattern recognition—happens automatically and quickly whereas analytical cognition—which is conscious deliberation—requires longer mental processing.

*Intuitive cognition* involves unconscious situational pattern recognition [50]. Intuitive cognition is independent of conscious “executive” control. It is large- capacity, fast, and it entails speeded judgments of situational patterns and no symbolic calculation. Intuitive cognition generally entails procedural memory and implicit knowledge (i.e., knowledge that cannot be consciously recollected). Intuitive cognition is ‘situated’—an individual can recognize a given situational pattern only when inhabiting the proper context [50]. Intuitive cognition is also strongly linked to emotion (e.g., ventromedial prefrontal cortex; [54]; [55]; [56]; [57]). The output of intuitive cognition’s processing—which is unconscious—is experienced consciously as a gut feeling [58].

*Analytical cognition* entails conscious deliberation that draws on limited working memory resources [59]; [60]; [61]). Analytical cognition is effortful, slow, and it involves the use of symbols, rules and algorithms. It includes declarative memory and explicit knowledge. Analytical cognition is ‘non-situated’—an individual can consciously deliberate across many situations, which allows for hypothetical thinking [50].

These two types of decision making—intuitive versus analytical—can be dissociated experimentally ([62], [63]) and neurologically ([64]; [65]; [54]; [66]; [55]). Moreover, Patterson and Eggleston [50] have shown that, across five very large literatures, intuitive cognition dominates responding in 80% or more of cases. This domination by intuitive cognition occurred in laboratory tasks as well as in real-world tasks.

As discussed below, it appears that both analytical cognition and intuitive cognition can engage in sensemaking, but they do so in different ways. We now turn to what we signify by the term sensemaking. Our conceptualization of sensemaking combines two theoretical perspectives (i.e., sign relations; frames).

### B. *Sensemaking*

The term sensemaking has been conceptualized variously as: (1) the process of encoding data in a given representation to answer task-specific questions ([67]); (2) constructing mental sense of one’s own world ([68]); or (3) mentally fitting data elements into contextual frames ([69],[70]). Yet we conceptualize sensemaking as involving the concepts of ‘sign relations’ and ‘frames’, as discussed next.

### C. Sign Relations

Sensemaking can be conceptualized as *sign interpretation*, which is called 'meaning making' in the field of semiotics (e.g., [71]; [72]; [73], [74]; [2]) a field originating from the writings of de Saussure ([75], [76]) and Peirce [77]. For example, Peirce [77] provided original analysis of the issues of thinking and meaning making. Peirce treated meaning making as sign interpretation—the meaning of a given thought occurs due to a triadic relation among the thought, the interpretation ("interpretant") of the thought as a sign (meaning), and a determining thought that the sign denotes [78].

In simpler terms, making sense of an object or event is found in its interpretation as a sign denoting some other (determining) object or event, which can be causal or correlational. Bains [71] discussed how sensemaking entails relations that function as signs—sign relations. The interpretation of such relations as signs is based on knowledge about situated patterns, which includes the grounding of objects or events to the real world ([79], [80]). For example, making sense of a traffic jam during a morning commute—due to an accident—would be found in its interpretation as a sign denoting that the person will be late for work.

### D. Frames

Sensemaking can also be conceptualized as involving 'frames', as suggested by Minsky ([81]; see also [82], [83], [84]); [85]; [86]). Minsky argued that when a new situation is encountered, a mental data-structure or mental model for representing stereotypical situations, called a frame, is retrieved from memory. The details of a given frame are adapted to fit with the current environmental context. Frames are also understood hierarchically. The context of a given situation is represented by higher, fixed levels of a frame whereas the specifics of the situation are represented by arguments or terminals that fit within the frame. According to Minsky [81], the relations among the lower-level terminals and their frame denote meaning. The concepts of scripts and plans, which are designed for representing sequences of events, are similar to frames [87]; [88]). Another concept closely allied to that of a frame is a schema (plural: schemata), which refers to a flexible mental data structure representing context-dependent relations among concepts in memory ([89]; [90]; [91]; [92], [93]; [94]).

### E. Signs Embedded Within/Across Frames

Recall that, in semiotics, making sense of an object or event entails relations that function as signs [71]. Such relations can be bound together in a frame. That is, within any given frame, a given element of the frame may serve as a sign denoting another element of the same frame (within-frame sign relation). For example, recall the commuter who is stuck in traffic during a morning commute due to an accident. Being stuck in traffic could activate a traffic-accident frame composed of elements like injuries, ambulance, debris, and stalled or stationary traffic. The presence of injuries would serve as a sign denoting the arrival of an ambulance, while the significant debris on the road would serve as a sign denoting stationary traffic. The ambulance itself could also serve as a sign denoting the possibility of stationary traffic. These would be within-frame sign relations.

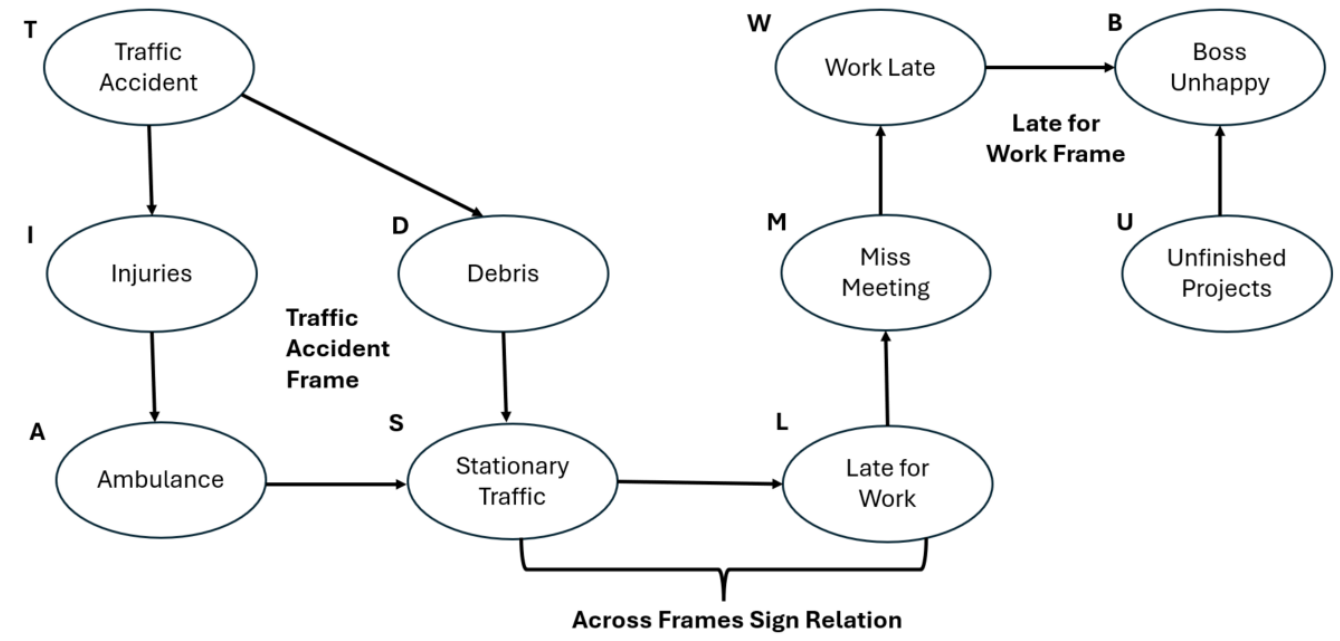

Fig. 1. Diagram depicting two frames, traffic-accident frame and late-for-work frame, and their within-frame sign relations and an across-frames sign relation. See text for detail.

Moreover, an element of a given frame may serve as a sign denoting an element of a different frame (across-frame sign relation). The stationary traffic of this traffic-accident frame would serve as a sign denoting that the commuter will be late for work, which would activate a new late-for-work frame. Thus, the stationary traffic (of the traffic-accident frame) and the late for work event (of the late-for-work frame) would comprise an across-frame sign relation. The late- for-work frame could be composed of elements such as missed meeting, work late, unfinished projects, and boss unhappy. The missed meeting could serve as a sign (its sense) denoting the need to work late, while the unfinished projects could serve as a sign (its sense) denoting an unhappy boss. The working late itself could also serve as a sign denoting the possibility of an unhappy boss. These latter relationships would be within-frame sign relations. See Figure 1.

Sensemaking also affords humans the opportunity to look back and interpret the potential causes of events. For example, the traffic-accident frame could have been preceded by a speeding- cars frame, which would have served as a sign denoting the traffic-accident frame; and the late- for-work frame could be followed by a possible-demotion frame, which would be denoted by the late-for-work frame. The sense derived from these within- and between-frame sign relations would allow humans to generate predictions about potential future events ([95], [96]), which could influence subsequent decision making.

Returning to dual processing, both analytical cognition and intuitive cognition can engage in sensemaking. Intuitive cognition (unconscious situational pattern recognition) can make sense of situational real-world patterns with appropriate experience. Thus, within a given frame, a given situational pattern could serve as a sign denoting another situational pattern or a certain action to undertake. Analytical cognition (conscious deliberation) can make sense of appropriate symbols and algorithms with appropriate training. Thus, within a given frame, a certain set of symbols could serve as a sign denoting another set of symbols or the solution to a particular problem.

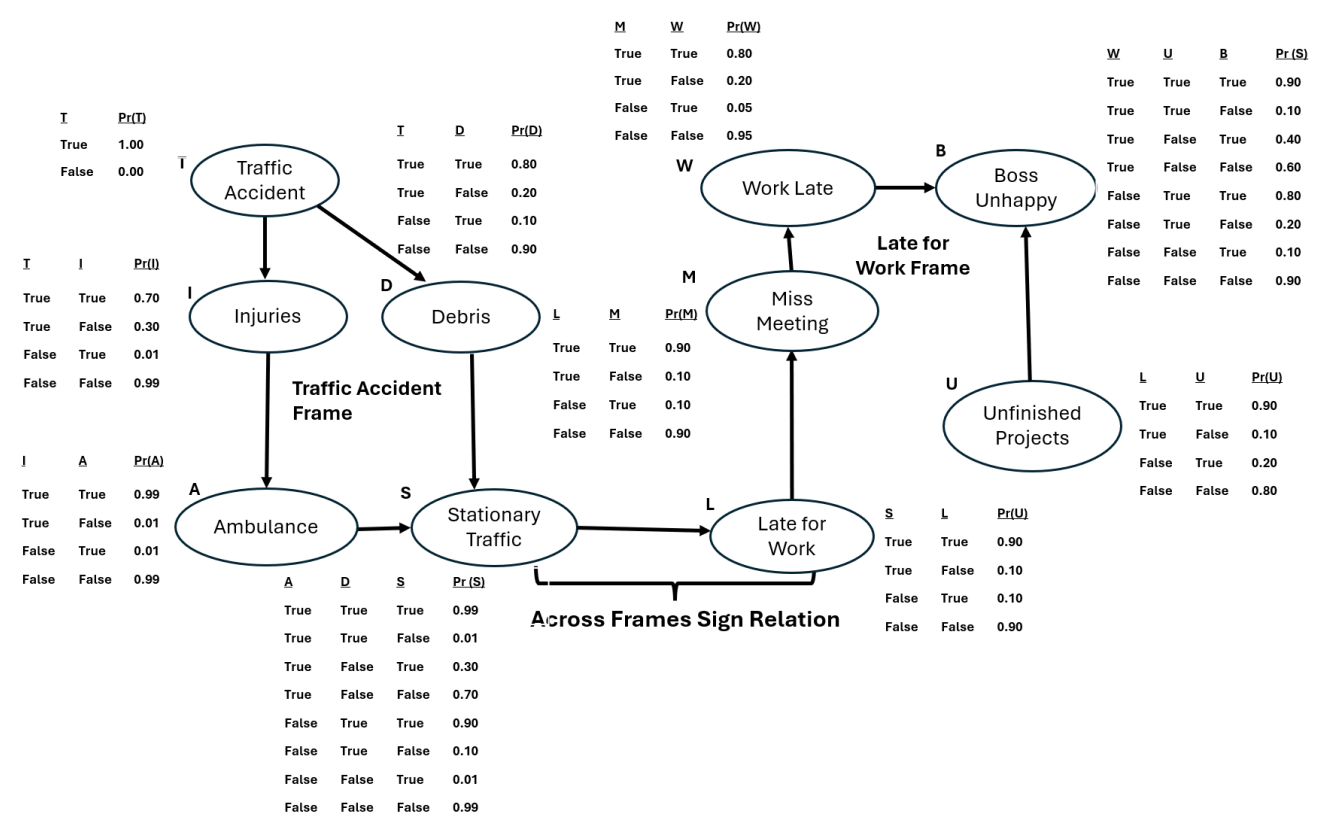

Fig. 2. Diagram depicting the two frames from Figure 1 shown as a Bayesian network with their conditional probability tables.

### F. Bayesian Network Modeling

A Bayesian network can be used to represent the unified framework of sensemaking, presented above. Bayesian networks (BN) are a form of probabilistic graphical model that employ a directed acyclic graph for encoding multi-dimensional probability distributions [97]. Bayesian networks represent information about a given domain using variables to symbolize propositions and directed edges to convey dependencies [98]. Probabilities are employed to quantify variables. Representing sensemaking and its attendant sign relations as probabilities in a BN reflects uncertainty [99], a natural feature of sensemaking in any realistic domain due to a lack of absolute knowledge. Such sign relations reflect conditional dependency.

Let us now return to the traffic-accident frame and late-for-work frame shown in Figure 1, but now represented as a BN with associated conditional probability tables (CPTs) as in Figure 2. The values in the CPTs are hypothetical estimates and are intended to show how a BN operates. We simulated the BN model shown in Figure 2 with a software tool called *Hugin* (version 8.9; [100]).

The prior probabilities of the model shown in Figure 2 are calculated as the following: T (traffic accident): 1.0; I (injuries): 0.7; A (ambulance): 0.7; D (debris): 0.8; S (stationary traffic): 0.81; L (late for work): 0.75; M (miss meeting): 0.75; W (work late): 0.61; U (unfinished projects): 0.73; and B (boss unhappy): 0.68. Thus, when the probability of debris is 80%, probability of being late for work is 75% and probability of boss being unhappy is 68%. The 80% probability of debris serves as a sign denoting that the commuter is more likely to be late than on time, the latter of which serves as a sign denoting that the boss is more likely to be unhappy than happy. When the probability of debris is made to be 0%, the posterior probability of being late for work is 27% and the posterior probability of the boss being unhappy is 41%. Here, the absence of debris serves as a sign denoting that the commuter is more likely to be on time than late, the latter of which serves as a sign denoting that the boss is more likely to be happy than unhappy.

### G. Sensemaking in AI Systems

Variants of sign relations and frames have been implemented in a variety of AI systems, each adopting different approaches to manage uncertainty, representation, and communication. Garcia et al. [101] illustrate a clear example of this in a two-agent system designed to create an emergent language for a categorization task. In this system, agents communicate by generating symbols that represent categories drawn from disentangled latent spaces, effectively producing what can be seen as sign relations. Each agent uses encoder-decoder networks to encode categorical information into symbols and decode them based on shared experiences. Through a reinforcement learning paradigm, the agents' interactions and symbol generation evolve to maximize mutual understanding, creating a self-organized sign language that balances between symbolic representations (signs) and probabilistic relations (sign relations) based on contextual categorization. In this setup, the experiment simulates the probabilistic and dynamic aspects of sign relations as described in human sensemaking, whereby symbols emerge to represent complex relationships across frames. (In the field of Semiotics, three types of signs have been delineated Peirce [77]: icon, which resembles its' object; index, which is factually connected to its' object; and symbol, which is connected to its' object by rule or convention. Because this distinction is not important to our effort, it will not be discussed further in this paper.)

Expanding on frames, this experimental design also hints at how frame-like structures might form in artificial systems. In the Garcia et al. [101] setup, each agent's latent space acts as an implicit frame that structures the relationship between signs and categories. The learned language reflects frame-like constraints as agents adapt their symbol generation and decoding strategies based on the probabilistic boundaries of these frames. This aligns with frame theory in that symbols and signs carry meaning not just individually but through a network of associations that reflect shared knowledge and categorical distinctions.

More broadly, representations such as large language models have been shown to have predictable implicit semantics ([26]; [102]), indicating the presence of implicit frame- like structures. Functional limitations in the flexibility of artificial cognition, however, suggest that the interplay between the analogues of sign relations and frames in these types of representations is not yet at the level of human cognition. Perhaps most glaringly, causal reasoning in machines continues to be much more difficult than statistical correlational reasoning [103].

## IV. PATTERN SYNTHESIS AND MEMORY RECOMBINATION

Sensemaking in novel situations may entail a form of pattern synthesis. Differences between a current novel situation and previous situations can be significant and sensemaking may require the synthesis of different attributes from different memories rather than retrieval of a single memory (Patterson & Eggleston, [50], [1], [146]; see also [69], [70]; [104]). This synthesized pattern could then be recognized as a sign in the sensemaking process (as well as be retrieved at a later time for further processing).

### A. Interactions Among Knowledge Structures

The generation of synthesized patterns as signs would require the interplay among various sources of knowledge. Based on this idea, knowledge structures would interact with each other to capture the generative capacity of human understanding in novel situations [105]. In humans, several types of memory have been identified [106]; [107]; [108]; [109], [110]): declarative memory (i.e., conscious recollection of facts and events); and nondeclarative memory (e.g., procedural memory—unconscious memory of invariant, relational knowledge supporting skill and behavioral dispositions; priming; classical conditioning). The generation of synthesized patterns as signs may entail the interaction among two or more of these memory systems.

### B. Insight and Sensemaking

Pattern synthesis and memory recombination during sensemaking is revealed by the insight literature. Insight refers to the sudden conscious realization of a problem solution (i.e., its sense) following a period of impasse, which typically occurs with nonroutine problem solving (e.g., [111]; [112]; [113]; [114]; [115]). Wallas [116] gave the original description of the phenomenon and suggested four stages of insight, which he called illumination: (1) preparation— conscious investigation of a problem; (2) incubation—unconscious processing; (3) illumination— "Aha" experience or sudden conscious insight due to previous unconscious processing during incubation; and (4) verification—conscious assessment of the insight. The first and fourth stages entail conscious processing whereas the second and third stages involve unconscious processing (e.g., [111]; [113]). Specifically, incubation is where the person recognizes the sense of a problem solution via intuitive cognition [50].

Insight problem solving is intuitive as shown by studies revealing that (1) cognitive processing leading to the problem solution during incubation was largely unconscious (e.g., [111]; [113]); (2) performance on insight problems was not linked with executive functions associated with working memory [117]; and (3) participants with impaired neurology in a region of cortex associated with working memory solved 50% more insight problems than healthy participants [118]. Disconnection from consciousness and working memory is a feature of intuitive cognition.

### C. Maier's (1931) Study

In a classic study of insight by Maier [113], which is discussed by Patterson and Eggleston [50], participants had to tie together ends of two long cords hung from the ceiling of a room and separated by a large distance. A solution was to make one cord a pendulum by tying an object (pliers) to its end and swinging it closer to the other cord so the person could grab both cords simultaneously. Some participants discovered the pendulum solution only after seeing the experimenter casually bump into and sway one of the cords while walking across the room ("help 1"). For 85% of the successful participants (with or without the help), the pendulum solution was discovered suddenly, from unconscious processing, and with insight ("Aha" experience) via intuitive cognition. For the successful participants, sensemaking of the pendulum would entail its interpretation as a sign denoting a solution to the two-cord problem.

The pendulum solution, as noted by Maier [113], was derived from a conceptual reorganization (pattern synthesis) of the weight of the pliers, length and position of the cord, and (for those who needed the help) the cord-swaying aspect of "help 1". Such synthesis of pattern to derive sense likely entailed memory recombination: knowledge about (1) how a length of cord behaves when weighted on one end, and (2) what a pair of pliers would weigh, would be retrieved and recombined from procedural memories (unconscious relational knowledge supporting skill development and behavioral tendencies tuned through experience; [109], [110]).

Memory recombination supporting intuitive situational pattern synthesis was fundamental to the insightful creation of the pendulum and its interpretation as a sign denoting a solution to the two-cord problem. Other studies on insight problem solving can be interpreted analogously (e.g., [111]).

### D. Compositional AI

The ability to recombine learned concepts, behaviors, and skills has long been recognized as a critical ability for AI systems (see, for example, the descriptions of necessary functionality in the original proposal on artificial intelligence by McCarthy, Minsky, Rochester, and Shannon [119]). Particularly in reinforcement learning, there is extensive literature on recombination of behaviors, including the influential work of Sutton et al. [120] on learning policies over low level options as well as more recent work on multi-task learning such as Andreas et al. [121]. In computer vision, many approaches have attempted to explicitly model the decomposition and recombination of conceptual primitives, for example, significant advances were made on deformable part-based object recognition [122]. It is crucial to note, however, that often presumed necessary explicit functionality is eventually shown to be less effective in machine systems than more simplistic methods that take better advantage of available computational capacity. Sutton [123] described this "bitter lesson" learned over the last 70 years of AI research. Although recent generative language models such as GPT-4 have shown improvements in the ability to summarize and combine disparate information in their context, this remains a difficult problem in general [26].

## V. DISTRIBUTED NETWORK REPRESENTATIONS

Cognitive processing frequently entails an interplay between various sources of knowledge in memory. For example, the insightful creation of the pendulum and its interpretation as a sign denoting a solution to the two-cord problem, discussed above, was derived from a memory recombination process supporting intuitive pattern synthesis [146]. Interactive processing, involving an interplay between bottom-up and top-down information, has been recognized in humans as being critical for behavior and also for consciousness ([124]; [125]; [126]). This interplay among different sources of knowledge is a feature of distributed representations [127].

### *A. Distributed Representation and Content Addressable Memory*

The existence of a distributed network representation assumes numerous highly interconnected units and no central processing center [128]. Theoretically, in one kind of distributed representation involving content addressable memory (CAM), each memory involves network units that have mutually excitatory interactions with units signifying each of its properties. Thus, activating an attribute of the memory would tend to activate the whole memory; and activating the whole memory would tend to activate all of its attributes. There would also be mutually inhibitory interactions between mutually incompatible attribute units. Thus, different memories would correspond to different patterns of activity over the same hardware units ([129]; [105]). One key feature of human memory is that it is content addressable—we can access information in memory based on most, if not all, attributes of the representation we are trying to retrieve [105]. The presence of distributed representations has been directly observed in human brain [130]; [131].

The results of the study by Maier [113] can be viewed in a distributed network representation scheme as shown in Figure 3. In Panel A, we have distributed representations of activated memories of a cord (e.g., with attributes strength, color, length, hung, knot on end) and of a pair of pliers (e.g., with attributes shape, weight, grip, color). As indicated in the figure, we hypothesize that the memory of the cord came from rope climbing in a gym class; and memory of the pliers came from carpentry work. In Panel B, we have the same representation as in Panel A except that we have added a new (non-activated) memory of a pendulum (which hypothetically came from seeing clocks) which has the attribute of 'cord sways' and also shares the attributes of 'weight', 'hung', and 'length' with the other memories. We call this a 'shared attribute' principle of distributed representations (see also [105]). In Panel B, we assume that the memory of the pendulum is not yet activated.

In Panel C, a new (activated) memory of the hint (which came from seeing the experimenter brush against the cord) is added, which shares the attribute of 'experimenter brushes cord' with the memory of the cord, and also shares the attribute of 'cord sways' with the memory of a pendulum. In Panel D, the pendulum memory has now become activated due to the presence of the hint and the activation of the attribute 'cord sways' being added to the coactivation of the attributes of 'length', 'hung', and 'weight'. In short, the memory of the hint interacts with the memories of the cord and the pliers to cause the dormant representation of the problem solution (pendulum) to become activated, and the participant has the 'Aha' (insight) experience[146].

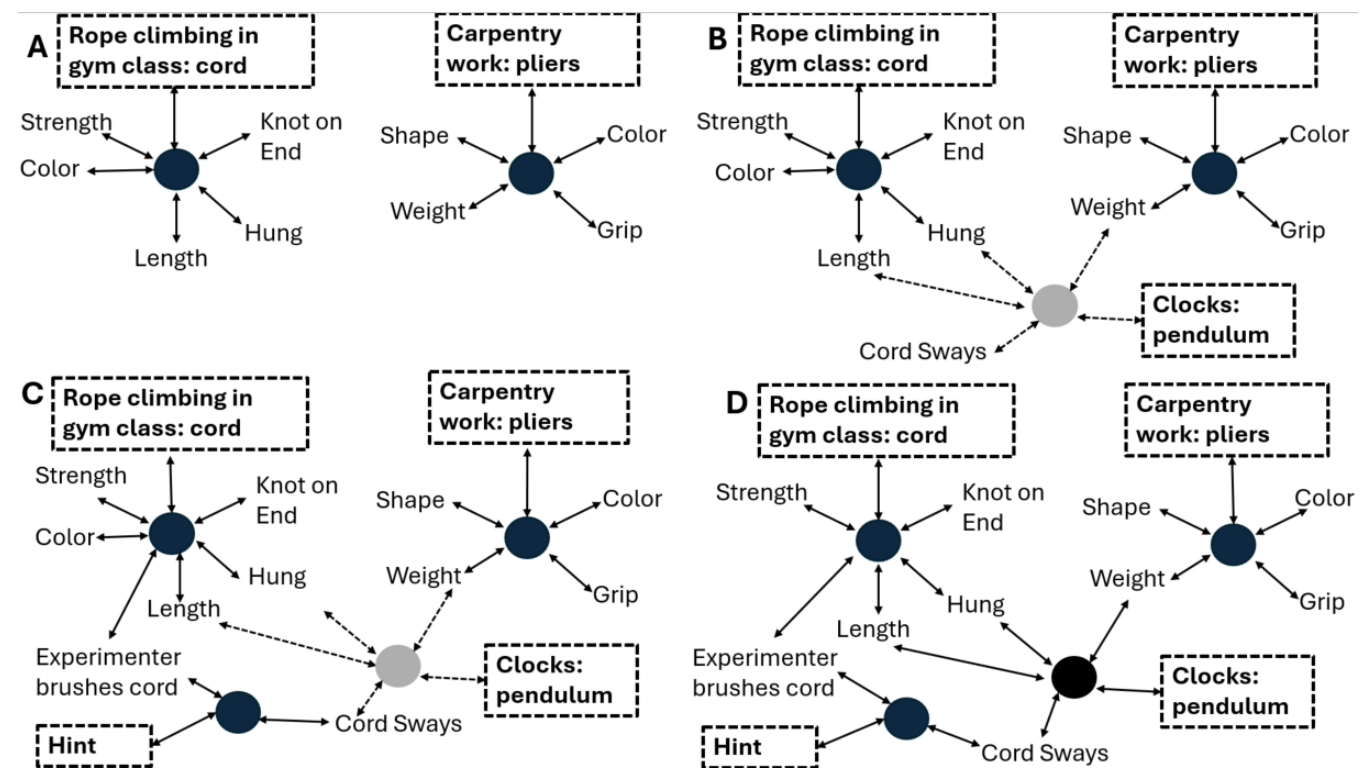

Fig. 3. Diagram depicting memories as a set of distributed network representations of items from a study by Maier [113]; figure is original. Panel A shows activated memories of a cord and of a pair of pliers. In Panel B, a new (nonactivated) memory of a pendulum has been added, which shares the attributes of 'weight', 'hung', and 'length' with the other memories. In Panel C, a new (activated) memory of the hint has been added, which shares the attribute of 'experimenter brushes cord' with the memory of the cord, and also shares the attribute of 'cord sways' with the memory of a pendulum. In Panel D, the pendulum memory has become activated, due to the presence of the hint and the activation of the attribute 'cord sways' being added to the coactivation of the attributes of 'length,' 'hung,' 'weight'.

### *B. Hypothetical Example*

Now consider a hypothetical example of sensemaking adopted from Patterson and Eggleston [1]. Imagine that an individual is on her way to work in the morning and stopped at a traffic light next to a gas station. Suddenly she sees a large dump truck speeding down a nearby steep hill. The truck is out of control and will likely strike the gas pumps. The site of the truck speeding toward the gas pumps would indicate that she could be injured by an explosion, and she would try to escape the situation. Yet how would she know about the danger of exploding gas pumps given that she has never experienced such an event in the past?

Because that exact event has never been previously experienced, the recognized danger could not be based on any single memory. Rather, the sight of the truck speeding toward the gas pumps would serve as a cue for the synthesis of a new frame that would entail a sign involving exploding gas pumps, derived from different memories from past experiences, and what the sign denoted, which would be the possibility of being injured. See Figure 4.

In synthesizing this new frame, there could be an episodic memory of not stopping one's car on a steep hill during a previous trip to San Francisco; another episodic memory of a collision involving a moving vehicle on an interstate (truck can't stop → collision, a within-frame sign relation); declarative memory of gasoline being a dangerous liquid during chemistry class; and an episodic memory of a friend being too close to exploding material and getting burned during a $4^{th}$ of July celebration (collision + gasoline danger → explosion, a within-frame sign relation). This synthesized element 'explosion' would serve as a new sign and its sense, or what that sign denoted, would be the possibility of injury (an across-frame sign relation). Thus, what we have is *sign relations produced by interaction among different representations of knowledge involving different kinds of memory*. See Figure 4. The idea of being injured would set up a new injury frame (not shown). Recognition of potential injury would be posted to consciousness as a feeling of fear.

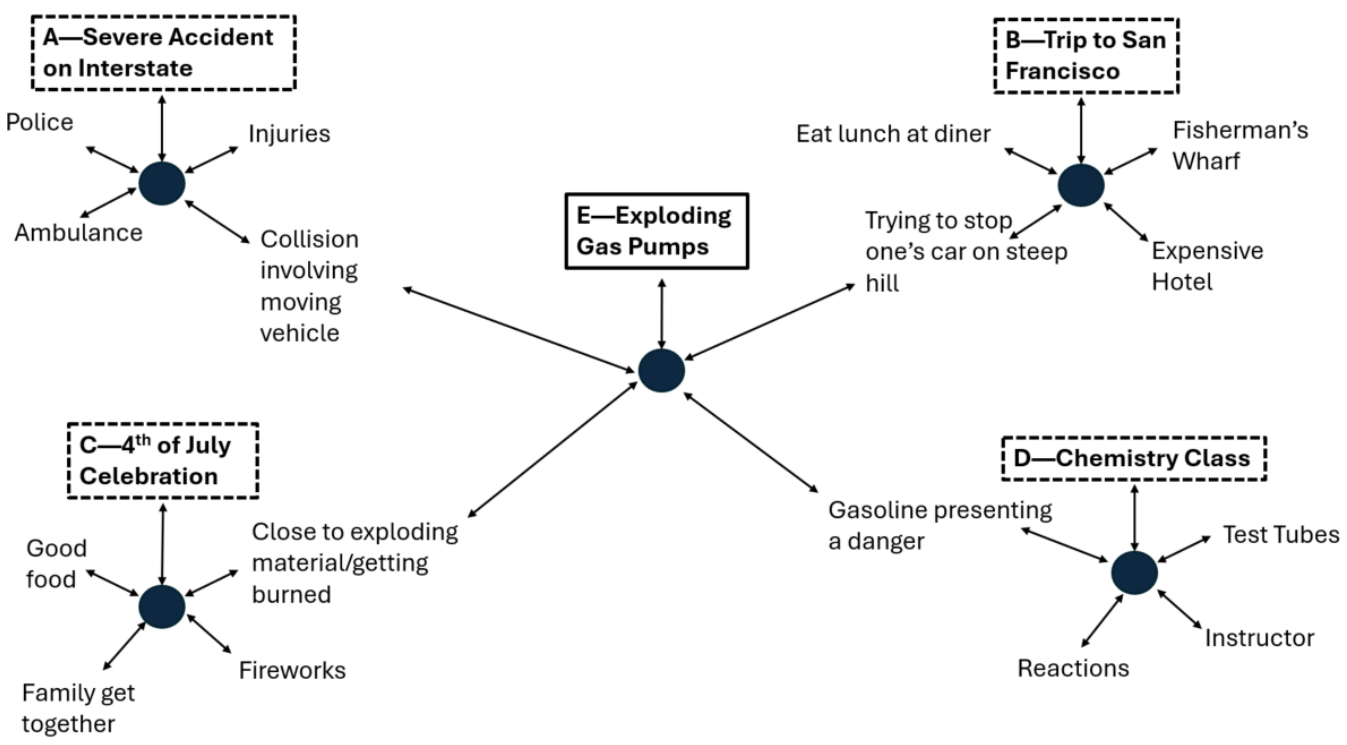

Fig. 4. Diagram depicting memories as a set of distributed network representations during sensemaking (sign interpretation) involving a hypothetical accidental explosion (adapted from Patterson & Eggleston, 2018). Assume that an individual is stopped at a traffic light next to a gas station and suddenly sees an out-of-control speeding dump truck that will likely strike the gas pumps. At the moment the out-of-control dump truck is sighted, different attributes from various memories—such as a collision involving a moving vehicle (panel A), trying to stop one's own car on a steep hill (panel B), being close to exploding material (panel C), and gasoline presenting a danger (panel D)—would be synthesized into a meaningful pattern, a sign, of exploding gas pumps (panel E). The sign of potential exploding gas pumps would denote possible injury (its sense) and the individual would immediately leave the situation Note that 'injury' could be a part of the Explosion frame if the person stayed at the scene and got injured. If the person immediately left the scene before the explosion, then injury would not be part of the frame and a new frame would be denoted, such as an 'Escaped explosion' frame (with potential embedded signs like 'hyperexcited', 'call police', and/or 'call family'.

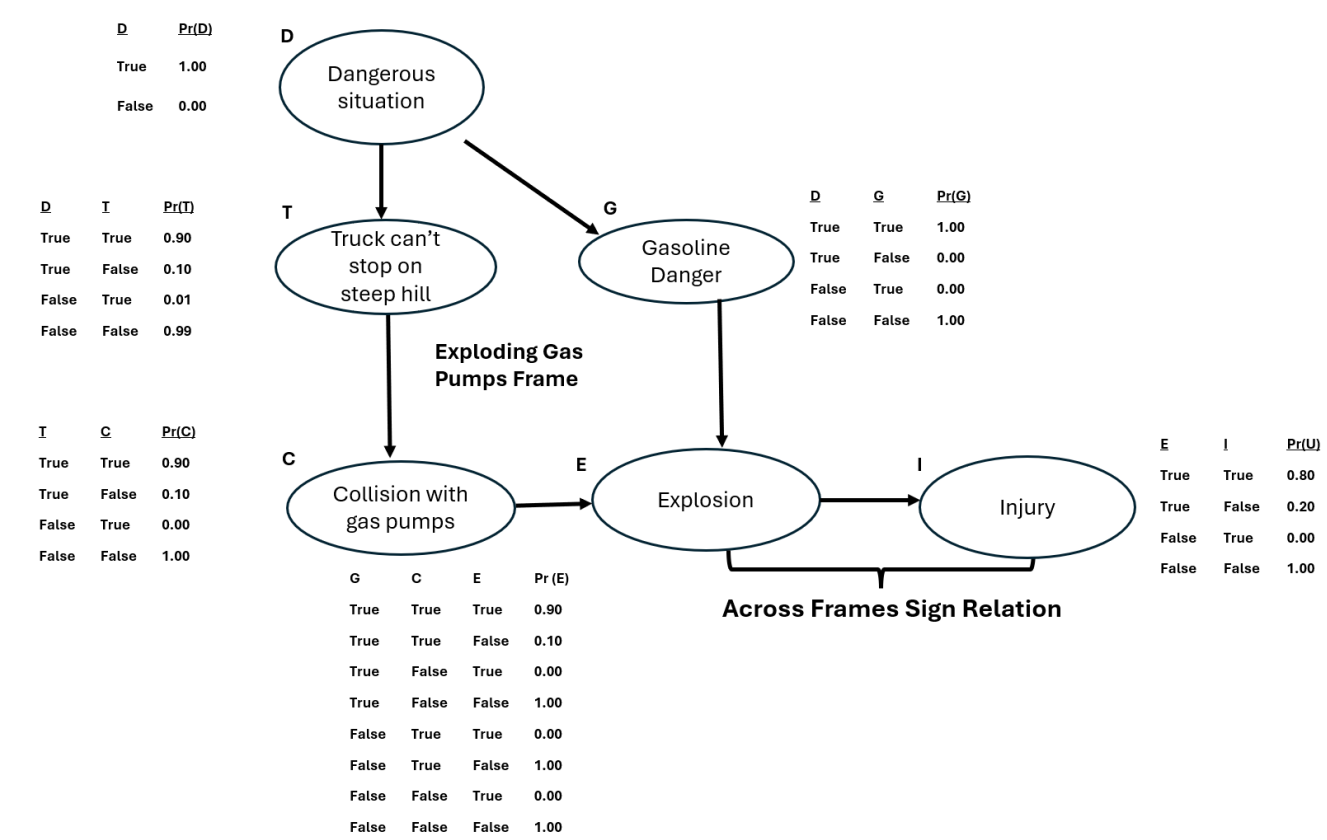

Fig. 5. Diagram depicting an exploding gas-pumps frame, synthesized from different memories. Hypothetically, an individual sees that a large dump truck is out of control as it speeds down a steep hill and will strike nearby gas pumps. The sight of the truck speeding toward the gas pumps would serve as a cue for the synthesis of a new frame with a sign involving an explosion. This synthesized frame and sign would be derived from aspects of the individual's various memories. The sense of this sign called 'explosion' would be the possibility of being injured.

We can represent this synthesized exploding-gas-pumps frame and attendant sign relations in a BN, similar to what was presented previously. The associated CPTs are given in Figure 5 (with the values hypothesized to come from previous memories), and this BN model was simulated with the software tool *Hugin* (version 8.9 [100]). In our simulation, the prior probability of an explosion is 0.73 and the prior probability of an injury is 0.58. However, if the probability of an explosion becomes 100% (i.e., the explosion variable is instantiated), then the posterior probability of an injury becomes 0.80.

In summary, the frame and accompanying sign relations used to inform sense making in humans can arise from a mental pattern-synthesis and memory-recombination operation carried out in novel environments. The benefit of this distributed network of knowledge for artificial agents is that it would allow for improved dynamic assessment of novel situations and better AI decision-making. This structure would afford AI more flexibility and ameliorate the problem of brittleness from supervised learning approaches, such as the problem of subtle superficial changes in image appearance leading to AI classification errors [31].

## C. Embodied Cognition

When sensemaking in novel situations, how does an individual determine which aspects of which memories are to be synthesized? We propose that each individual engages in a set of unconscious, embodied, intuitive simulations that attempt to align different combinations of various attributes from different memories until a solution to the sense of a given object, event, or situation is found; that sense is posted to consciousness as a certain gut feeling. With embodied cognition, the same neural substrate that underlies perception and action also underlies central cognitive processing ([132],[133], [134]; see also [135]). Embodied cognitive processes can be conceptualized as high-level "simulators" operating within modal systems of representation ([132], [133]). Patterson and Eggleston [1] have argued that intuitive cognition is embodied.

This idea of unconscious embodied intuitive simulation stands in opposition to standard cognitive theories based on the concept that representations in modal systems (e.g., vision) are transduced into amodal symbols for representing knowledge. In an amodal cognitive system, perceptual and motor systems would not be playing any significant role in "central" cognitive processing ([132]; [136]). However, the problem with amodal cognitive systems is that their symbols are arbitrary and lack grounding in the natural world—and grounding may be a necessary precondition for sensemaking ([79], [80]). A solution to the symbol grounding problem is to make cognition embodied [134].

## D. Actual Example

Now consider the following actual example that involved measuring human sensemaking using the approach that we are advocating in the present paper. To determine whether certain human-machine systems improve the sensemaking capability of intelligence analysts, Frame, Maresca, Christensen-Salem and Patterson [137] investigated the advantages of *Project Maven*, a machine learning-based recognition aid developed for the US Department of Defense. Project Maven automates the process of analyzing drone surveillance video [136, 138] by classifying objects, such as vehicles or people, in a scene [138, 139]. Frame et al. tested the idea that, by embracing detection and identification, Project Maven may alleviate the cognitive workload of analysts so that their sensemaking would be enhanced. Frame et al. defined sensemaking as sign relations embedded within and across frames.

Frame et al. [137] measured sensemaking by creating a set of 10 simulated intelligence, surveillance and reconnaissance (ISR) compound overwatch scenarios (3-minutes each) viewed as full-motion-video. Each scenario depicted a plot in a terrorist narrative that served as a sensemaking frame (e.g. hostage taking). Sensemaking was measured by cue identification and event prediction—that is, by determining how well the participants (n = 73) could: (1) predict the final ending of each of the 10 scenarios (‘prediction’) after seeing its beginning; and (2) identify the initial sign that denoted the final ending of each scenario (‘identification’).

Five overwatch scenarios contained signs that were people and/or vehicles and highlighted/tracked by simulated Maven (by placing a small globe symbol above them): IED explosion (sign: digging by road); people attacking hospital (sign: someone stealing ambulance); school bus explosion (sign: someone tampering with school bus); attack on first responders (sign: slum lit on fire); and running into protesters (sign: someone stealing car). Five overwatch scenarios contained signs that were not people/vehicles and not highlighted/tracked by simulated Maven: hostage taking (sign: rope); quick change evasion (sign: laundry/clothing); blowing up building (sign: gas canisters); taking over ambulance (sign: injured bodies); shooting at market (sign: dead drop gun). The results showed that having Project Maven direct the participants’ attention to people or vehicles serving as a sign did improve the participants’ ability to predict the final ending of the scenarios and thus enhance sensemaking.

## VI. Discussion

The ideas and observations expressed in the present paper offer information toward creating artificially-intelligent agents with the capacity for sensemaking in novel environments. With respect to sensemaking, our approach entails a unified framework for sensemaking that posits sign relations embedded within and across frames. That is, an element of a given frame may serve as a sign denoting an element of the same frame—i.e., within-frame sign relation. Or an element of a given frame may serve as a sign denoting an element of a different frame—i.e., across-frame sign relation. This unified framework of sensemaking is represented as a Bayesian network, which reflects uncertainty in the sensemaking process.

With regard to novel environments, our approach involves the idea of creating interaction among various distributed-knowledge structures. This type of memory representation is a form of CAM. Such interaction would be mediated via shared attributes among the various memories [138]. Thus, activating a memory would tend to activate all of its attributes, including any shared attributes. Activation of those shared attributes would tend to activate other memories that also share those attributes (there would also be mutually inhibitory interactions between mutually incompatible attribute units). For those activated memories agreeing on an attribute, the node for that attribute would be strongly activated, which would tend to activate other representations. And for those activated memories not agreeing on an attribute, the different attributes would tend to cancel each other out and be suppressed. The net response of the ensemble of network units would be a *synthesized representation of some object, event, or situation that would serve as a unique sign for sensemaking in a novel environment*. This scheme involving synthesized representations serving as signs provides an innovative and original framework for sensemaking in novel environments.

In sum, aspects and attributes across many memories can be shared and recombined in unique ways to create synthesized signs which then denote certain outcomes. In this way, the frame and accompanying sign relations used to define sensemaking can arise from a mental pattern- synthesis and memory-recombination operation carried out in novel environments. Such synthesized signs can be produced by distributed representations of interacting memories that share attributes. The interplay among different sources of knowledge in memory is a key feature of distributed representations underlying human cognitive processing.

### A. Markov Decision Processes

Currently, the ability to make sense in novel environments is outside the scope of contemporary computing science. For instance, consider the field of reinforcement learning (RL). RL refers to a computational approach for automating goal-directed learning and decision making that comes from the direct interaction with the environment. RL uses the formal framework of Markov decision processes (MDPs) which entails the interaction between a learning agent and its environment in terms of states, actions, and rewards. This type of trial-and-error-learning involves learning how to map situations onto actions so as to maximize a reward signal. The concept of trial-and-error learning emerged early in the field of artificial intelligence ([139]; [140]; [141]).

With MDPs, the probability of each possible value for the state and reward depends only on the immediately preceding state and action and not at all on earlier states and actions. This is called the Markov property. With the Markov property, the conditional probability distribution of future states depends only upon the present state; given the present, the future does not depend on the past. This means that the information from past memories would not be available for any pattern-synthesis and memory-recombination operation for sensemaking in novel environments.

### B. Concluding Remarks

To implement sign relations and frames in an AI system, we can follow the approach taken by Garcia et al. [101], who designed a two-agent system for creating an emergent language for a categorization task. In this approach, agents communicate by generating symbols that represent categories drawn from disentangled latent spaces, mapping each latent factor to a generative factor, which effectively produces sign relations. Through reinforcement learning, agents’ interactions and symbol generation evolve to maximize mutual understanding, which creates a self-organized sign language involving symbolic representations (i.e., signs) and probabilistic relations (i.e., sign relations) based on contextual categorization.

## APPENDIX

Appendix I: Algorithm Description (proposed, 1)

**Algorithm 1: Sensemaking Using Bayesian Networks and Constraint Satisfaction Networks**

1. **Initialize Bayesian Network (BN):**
   - Define nodes that represent **events and signs** within frames, incorporating elements that denote causal or contextual information within and across frames.
     - **Example Nodes**: Traffic_Accident, Explosion, Injury.
   - Establish edges representing **sign relations** (intra-frame) and **causal links** (across frames) to signify dependencies between signs in different frames.
     - **Example Links**: Traffic_Accident → Explosion, Explosion → Injury
2. **Assign Conditional Probability Distributions (CPDs):**
   - Assign probabilities to each node using expert knowledge or hypothetical data for illustration, as noted in the paper's hypothetical examples.
   - Incorporate probabilities that denote **the impact of certain events on expected outcomes**, making sense of the likely signs and outcomes within each frame.
     - **Example**:
       * P(Traffic_Accident)=[0.7,0.3]
       * P(Explosion—Traffic_Accident)=[0.8,0.2] if Traffic_Accident = 1, and [0.3,0.7] if Traffic_Accident = 0.
       * P(Injury—Explosion)=[0.9,0.1]P(Injury | Explosion) = [0.9, 0.1]P(Injury—Explosion)=[0.9,0.1] if Explosion = 1, and [0.4,0.6] if Explosion = 0.
3. **Add CPDs to the Network:**
   - Incorporate all CPDs into the BN model, verifying that the model's structure aligns with the described interdependencies between events and their probabilities.
4. **Perform Initial Inference on BN:**
   - Use variable elimination or another inference method to assess potential outcomes given current evidence, noting how the evidence impacts probabilistic outcomes within and across frames.
     - **Query Example**: P(Injury—Traffic_Accident=1)
5. **Initialize Constraint Satisfaction Network (CSN):**
   - Define **distributed memory nodes**, where each node represents attributes linked by shared signs. These signs activate across related frames, supporting inter-frame synthesis.
   - Define constraints that manage **excitatory and inhibitory interactions** between shared attributes, reconciling conflicting or overlapping memory attributes across frames.
     - **Example Constraint**: Debris and ClearPath are mutually exclusive attributes that cannot coexist without triggering a conflict.

6. **Add Memory Nodes to CSN:**
    - Assign attributes to each memory node, using **observed signs and shared characteristics across frames**. These attributes help synthesize memory nodes based on overlapping experiences.
        - **Example**:
            * Memory1 = {Debris: 1, ClearPath: 0}
            * Memory2 = {Noise: 1, Accident: 0}.
7. **Resolve Constraints in CSN:**
    - Apply **constraint satisfaction through relaxation and synthesis** to resolve memory attributes iteratively until reaching a stable state, where memory nodes reflect a unified representation of potential outcomes.
    - During this resolution, prioritize adjustments based on **strongest constraints** that guide memory activation to support the most coherent and stable interpretation.
        - **Example Adjustment**: Resolve conflicts between Memory1 and Memory2 by aligning compatible states and suppressing incompatible ones
8. **Update BN with Resolved Evidence:**
    - Feed evidence from the CSN into the Bayesian network, updating node states or CPDs based on the synthesized outcome. This evidence reflects the final attribute synthesis in the CSN, adjusted to the BN's probabilistic framework.
        - **Example Update**: If conflict resolution suggests Explosion = 0, adjust BN inferences accordingly.
9. **Final Inference on BN:**
    - Conduct a final inference to predict outcomes based on the CSN-updated evidence, generating posterior probabilities for the most likely outcomes based on reconciled frames.
        - **Query Example**: P(Injury—Traffic_Accident=1,Explosion=0)
10. **Return Sensemaking Decision (Feedback Loop):**
    - Use the results of the final inference to formulate a sensemaking recommendation. Continue this feedback loop as necessary, updating both CSN and BN iteratively with new information and reconciled conflicts.
    - **Example**: Suggest a detour if the probability of Injury is high given current evidence.